\documentclass[times,twocolumn,final,numbers, sort&compress,square]{elsarticle}

\usepackage{ycviu}
\usepackage{framed,multirow}

\usepackage{amssymb}
\usepackage{latexsym}

\usepackage[hidelinks]{hyperref}
\usepackage{url}
\usepackage{xcolor}
\definecolor{newcolor}{rgb}{.8,.349,.1}

\usepackage{epsfig}
\usepackage{graphicx}
\usepackage{amsmath}
\usepackage{amssymb}
\usepackage{multirow}
\usepackage{tabulary}
\usepackage{tabularx}
\usepackage[tight,normalsize,sf,SF]{subfigure}
\usepackage{color}
\usepackage{cite}
\usepackage{soul}
\usepackage[inline]{enumitem}
\usepackage{hhline}
\usepackage{algorithm}
\usepackage{algpseudocode}
\usepackage{array}
\usepackage{flushend}
\usepackage{multirow}
\usepackage{tabulary}
\usepackage{tabularx}
\usepackage{bm}
\usepackage[tight,normalsize,sf,SF]{subfigure}
\usepackage{color}
\usepackage{csquotes}

\DeclareMathSymbol{\Lambda}{\mathalpha}{operators}{3}
\DeclareMathSymbol{\Pi}{\mathalpha}{operators}{5}
\newcolumntype{P}[1]{>{\centering\arraybackslash}p{#1}}
\graphicspath{{./figs/}}
\DeclareGraphicsExtensions{.eps,.ps,.pdf,.png,.tiff}

\usepackage{array}
\newcolumntype{P}[1]{>{\centering\arraybackslash}p{#1}}
\newcolumntype{M}[1]{>{\centering\arraybackslash}m{#1}}

\begin{document}

\begin{frontmatter}

\title{Prediction and Description of Near-Future Activities in Video}

\author[1]{Tahmida Mahmud} 
\author[1]{Mohammad Billah}
\author[2]{Mahmudul Hasan}
\author[1]{Amit K. Roy-Chowdhury}


\address[1]{Department of Electrical and Computer Engineering, University of California, Riverside, CA 92521, USA}
\address[2]{Comcast AI, Washington, DC-20005, USA}

\begin{abstract}
Most of the existing works on human activity analysis focus on recognition or early recognition of the activity labels from complete or partial observations. Similarly, almost all of the existing video captioning approaches focus on the observed events in videos. Predicting the labels and the captions of future activities where no frames of the predicted activities have been observed is a challenging problem, with important applications that require anticipatory response. In this work, we propose a system that can infer the labels and the captions of a sequence of future activities. Our proposed network for label prediction of a future activity sequence has three branches where the first branch takes visual features from the objects present in the scene, the second branch takes observed sequential activity features, and the third branch captures the last observed activity features. The predicted labels and the observed scene context are then mapped to meaningful captions using a sequence-to-sequence learning-based method. Experiments on four challenging activity analysis datasets and a video description dataset demonstrate that our label prediction approach achieves comparable performance with the state-of-the-arts and our captioning framework outperform the state-of-the-arts.
\end{abstract}

\end{frontmatter}


\section{Introduction}\label{sec:intro}
%
%
%
%
\footnotetext[1]{The paper has been accepted for publication in Computer Vision and Image Understanding.}

Activity analysis is a widely studied problem in the computer vision community. A large number of the existing works focus on recognition of observed activities or early recognition of partially observed activities. Predicting the labels of future activities which have not yet been observed is different from the recognition problem, where inferences need to be made on activity features which have been observed. The word `prediction' has been used in \citep{cao2016activity, li2014prediction, li2016recognition, ryoo2011human, wang2016context}, referring to the early recognition task, i.e., predicting the label of the ongoing activity where the first few frames have already been observed. However, in the prediction problem we are addressing, \emph{no observation is available beforehand}. Predicting the future activity labels is critical in real life scenarios, where anticipatory response is required based on an observed segment of the video, e.g., driver intent prediction \citep{zyner2017long, morris2011lane} in Advanced Driver Assistance Systems (ADAS) where a description of which lane the driver might move into in the near future is necessary to predict the likelihood of potential collisions in complex traffic scenarios, or Human Intent Prediction (HIP) \citep{mcghan2015human, townsend2017estimating} in human-robot collaboration where the robot may need to predict what the human may do in the future to ensure safety and efficiency. There are a number of approaches \citep{chakraborty2014context,kitani2012activity,mahmud2017joint,abu2018will, mehrasa2019variational,sun2019relational,abu2019uncertainty,furnari2020rolling,gammulle2019forecasting,gammulle2019predicting,ke2019time,liang2019peeking,miech2019leveraging} which perform label prediction on real-life activity datasets. 

\begin{figure*}[t]
	\centering
	\includegraphics[width=0.7\linewidth] {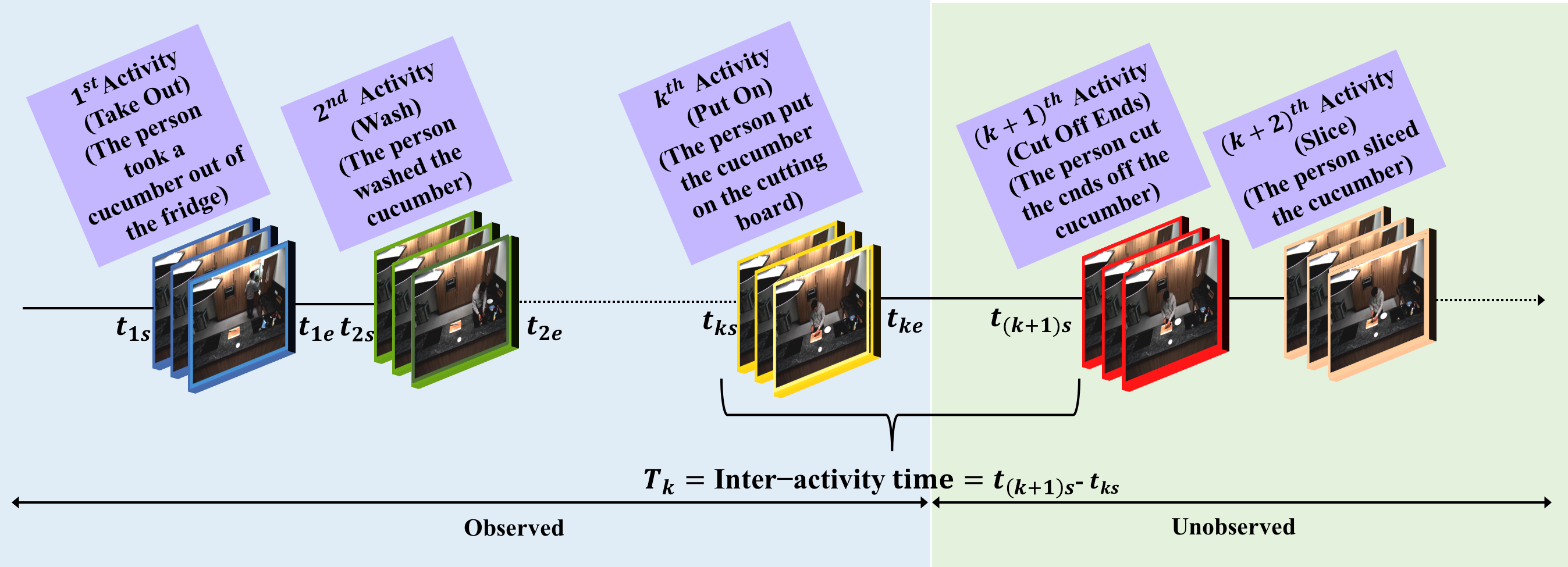}
	\caption[test]{There are $k$ activities in the observed portion of a video with starting times ($t_{1s}$, $t_{2s}$, ..., $t_{ks}$) and ending times ($t_{1e}$, $t_{2e}$, ..., $t_{ke}$). We want to predict the labels and the captions of $(k+1)^{th}$, $(k+2)^{th}$, ... activities. \footnotemark[1] }
	\label{fig:fig1a}
\end{figure*}

Generating description of visual content is an active research area in both computer vision and natural language processing community. Since vision and language are two of the richest interaction modalities available to humans, it is crucial to understand the relationship between them. Language is the most natural way to make information from any semantic representation meaningful. In the last few years, this problem has received significant attention for image captioning \citep{farhadi2010every, kulkarni2013babytalk, vinyals2015show, johnson2016densecap} as well as video captioning \citep{das2013thousand, guadarrama2013youtube2text, krishnamoorthy2013generating, donahue2015long, venugopalan2015sequence, venugopalan2014translating, yao2015describing, barbu2012video, khan2012describing, rohrbach2013translating, yu2016video, krishna2017dense, rohrbach2015long, liang2017recurrent, shin2016beyond, yu2018fine, xiong2018move}. Unlike image description, video description has to deal not only with the appearance of the objects but also with motion over time. There has been significant work in the multimedia community on joint image and text processing, e.g., for retrieval, captioning etc. \citep{shetty2016frame,xu2017learning,wang2019cross,chowdhury2018webly,yan2019stat,gao2017video,dong2018predicting,sener2019zero}; however, none of them except \citep{sener2019zero} address the problem of predicting captions for the near-future events in videos. To the best of our knowledge, almost all of the existing works on video captioning focus on the observed portion of the video, i.e., describe events which have already happened or happening at the moment. Ours is one of the earliest works where we look into the problem of providing captions for a sequence of \emph{near-future unobserved events in videos}. Generating the labels of future unobserved activities can be considered as the first step towards describing the future. But it may be desirable to offer a richer description than a simple one-word/phrase label for specific applications like assistive systems \citep{elmannai2017sensor, owayjan2015smart} for the visually impaired. There has been work on generating future frames \citep{vondrick2015anticipating}, which are much richer in content, but the approach is constrained to only a few such frames. Our work lies in between these two extremes: it can generate semantically meaningful captions that describe changes in activities and thus able to predict much further in time than the frame generation work \citep{vondrick2015anticipating}, while at the same time, provides a much richer description than label prediction \citep{chakraborty2014context, kitani2012activity, mahmud2017joint, abu2018will}. 


\subsection{Problem Definition}
For a video observed up to a certain time, we want to predict the labels of the future activity sequence and provide a caption describing these future activities in the context of the observed video. This is illustrated in Fig.~\ref{fig:fig1a}. Here, we have only observed up to $k^{th}$ activity. Now we want to predict the labels and the captions of the future activity sequence, i.e., the labels and the captions of the $(k+1)^{th}$, $(k+2)^{th}$, $\cdots$ activities and the starting time of that sequence, i.e., $t_{(k+1)s}$. In our experiments, each predicted sequence contains three unobserved future events (chosen empirically because predicting beyond three activities is very error prone).

\begin{figure*}[t]
	\centering
	\includegraphics[width=0.65\linewidth] {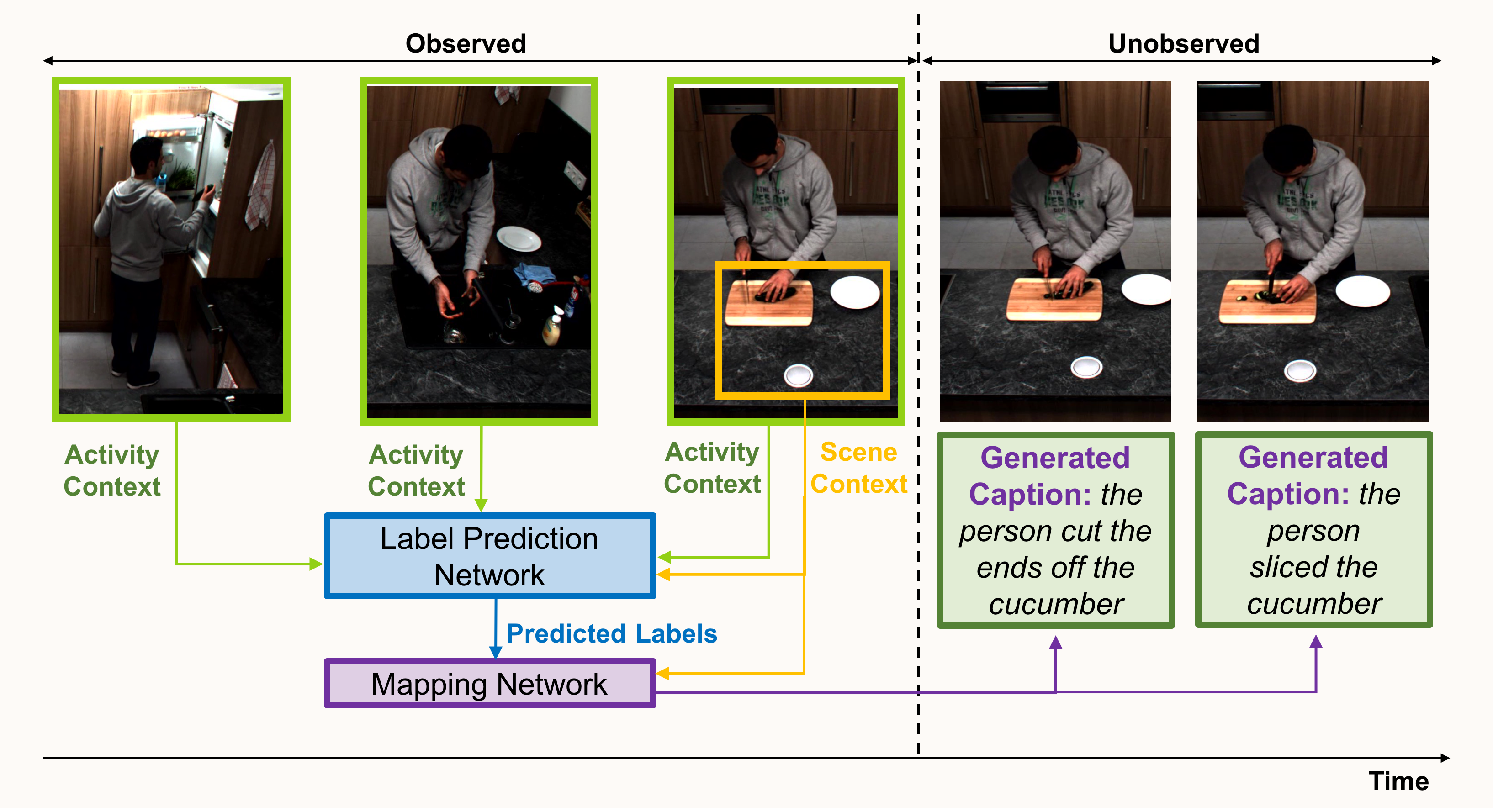}
	\caption{Overview of our approach. The label prediction network is trained on both the sequential activity features from previously observed activities and the object features present in the last observed portion of the scene. The sequence-to-sequence learning-based mapping network finally maps the sequential labels and observed scene context to a sequence of captions. A detailed version of this figure is given in Fig.~\ref{fig:fig4}.}
	\label{fig:fig2}
\end{figure*}

\subsection{Overview of the Approach}
In this paper, we present an integrated approach to answer two important questions regarding the unobserved portion of a video observed up to a particular time: {\it what activities will happen next}, and  {\it what captions describe them best}. We predict the {\bf labels} of a sequence of future unobserved activities in both coarse (VIRAT Ground Dataset \citep{oh2011large}) and fine grained activity datasets (MPII-Cooking Dataset \citep{rohrbach2012database}, MPII-Cooking $2$ Dataset \citep{rohrbach2016recognizing} and EPIC-KITCHENS Dataset \citep{Damen2018EPICKITCHENS}). This is posed as a joint label and starting time prediction task because intuitively the problem of predicting the label and the starting time of unobserved activities are closely related. For example, in MPII-Cooking Dataset, `cut slices' can be followed by two probable activities: `spice' or `take out from drawer'. Usually, `spice' takes place immediately after `cut slices'; but if there is a delay, then `take out from drawer' happens before. Once the labels are available, we map them along with the scene context of the last observed portion to generate meaningful {\bf captions} for a sequence of future activities. Instead of using a rule- or template-based natural language generation approach, we are motivated by the data driven domain-independent learning-based approach \citep{sutskever2014sequence} which replaced rule based methods in statistical machine translation. Instead of performing the mapping between two language spaces, we are doing a mapping from labels to captions. This sequence-to-sequence learning-based approach as used in other captioning works \citep{donahue2015long, rohrbach2015long} makes minimal assumptions on the sequence structure. 

\footnotetext[1]{Please note that although the captions are generated for future unobserved events, they are in past tense since the reference human descriptions are given in past tense in the dataset which we used during training to generate coherent results for meaningful comparison with \citep{rohrbach2014coherent} and \citep{yu2016video}.}

An overview of our proposed framework is illustrated in Fig.~\ref{fig:fig2}. To infer the future unobserved activity sequence, we use context information (relationship between actors, objects, and activities) from the observed portion of the videos. We incorporate sequential activity context (temporal ordering among activities), scene context (objects present in the last observed portion of the scene) and inter-activity time (difference between the starting time of the observed activity and the future activity) context. The details of the context information are provided in Section \ref{section:Context1}. We develop a deep network for label prediction by merging three branches: one with two fully connected layers, another with two LSTM layers and the last one with another two fully connected layers. The network is trained on the previous activity features and the features of the objects present in the last observed portion of the scene. For captioning a sequence of future activities, we use a multi-layered LSTM to map the predicted labels and observed scene context to a fixed dimensional vector.  Another deep LSTM conditioned on the input sequence is used for extracting the target caption from that vector. The ability of LSTM layers to incorporate long term sequential dependencies makes it a suitable choice for this application. 

\subsection{Main Contributions}
Our work focuses on providing a description of what activities may happen in the near-future from a sequence of current observations. We first predict the labels of the future activities, which is then followed by captioning. To the best of our knowledge, almost all of the existing works on video captioning focus on the observed portion of the video, and ours is one of the earliest works for captioning a sequence of near-future unobserved events in videos. It provides a richer description than just predicting the label of the next activity and predicts over a longer duration than works on frame prediction. The {\bf main contributions} of this work are: 
\begin{enumerate} 
	\item{We jointly model the sequential relationships of the activities, scene context and the last observed activity features in order to predict the labels of a future activity sequence.} 
	\item{We solve a novel and relevant problem of captioning a sequence of future unobserved events of a video using a sequence-to-sequence learning-based approach.}
	\item{We perform extensive experiments that show the effectiveness of the proposed framework.}
\end{enumerate}

\subsection{Organization of the Paper}
The rest of the paper is organized as follows. Section \ref{section:related} gives a description of the state-of-the-art approaches in applications of LSTMs, activity analysis, and video captioning. The role of different context attributes is provided in Section \ref{section:Context1} and Section \ref{section:Context2}. The network architectures are provided in Section \ref{section:Architecture1} and Section \ref{section:Architecture2}  and model training details are provided in Section \ref{section:Training1} and Section \ref{section:Training2}. A test case scenario is explained in Section \ref{section:Test}. Experimental results and comparisons with state-of-the-art methods are shown in Section \ref{section:Experiments} and conclusions are drawn in Section \ref{section:Conclusion}.

\section{Relation to Existing Works}
\label{section:related}
Our work involves the following areas of interest: Long Short-Term Memory (LSTM) network, future activity label prediction, and video captioning. We will review some relevant papers from these areas.

{\bf Long Short-Term Memory (LSTM) Network.}
Unlike traditional neural networks, Recurrent Neural Network (RNN) has the capability of allowing information to be passed from one step of the network to the next using the loops inherent to their structure. However, in practice, RNNs cannot handle long-term dependencies, primarily because of the vanishing and exploding gradient problem. To overcome the challenge of handling long-term dependency, a special type of RNN called LSTM (Long Short-Term Memory) was introduced in \citep{hochreiter1997long}. LSTMs have achieved impressive performance in different sequence learning problems \citep{donahue2015long,graves2014towards,pinheiro2014recurrent,sutskever2014sequence,visin2016reseg}. Its ability to capture long-range dependencies makes it a perfect tool for long-term context incorporation.

{\bf Future Activity Label Prediction.} There have been a few works which predict the future unobserved activity such as approaches using Hierarchical state space Markov Chains \citep{lade2010task}, semantic scene labeling \citep{kitani2012activity}, Probabilistic Suffix Tree (PST) \citep{li2014prediction}, augmented- Hidden Conditional Random Field (a-HCRF) \citep{wei2014forecasting}, Markov Random Field (MRF) \citep{chakraborty2014context}, kernel-based reinforcement learning \citep{huang2014action}, max-margin learning \citep{lan2014hierarchical}, and deep network \citep{vondrick2015anticipating, abu2018will, rhinehart2017first, mahmud2017joint,mehrasa2019variational,sun2019relational,abu2019uncertainty,furnari2020rolling,gammulle2019forecasting,gammulle2019predicting,ke2019time,liang2019peeking,miech2019leveraging}. Among these, \citep{chakraborty2014context,kitani2012activity, abu2018will, mahmud2017joint, mehrasa2019variational,sun2019relational,abu2019uncertainty,furnari2020rolling,gammulle2019forecasting,gammulle2019predicting,ke2019time,liang2019peeking,miech2019leveraging} perform prediction, without any observation of the activity to be predicted, in the label space. In \citep{vondrick2015anticipating}, where visual representation of images is predicted and then recognition algorithm is applied, actions can be anticipated only up to one second in the future. The focus of \citep{rhinehart2017first} is forecasting behavior/goal where the fundamental state variables involved are different than the label space. \citep{abu2018will} infers about the labels of a future activity sequence using a CNN-based and a RNN-based approach. However, they predict the labels of a future unobserved activity sequence only; whereas the main focus of this work is predicting the captions of a future activity sequence. \emph{Our previous work on activity prediction \citep{mahmud2017joint} has achieved the highest accuracy on two challenging activity datasets incorporating different context attributes but did not perform sequence prediction.}

{\bf Video Captioning.}
The initial works on video captioning \citep{kojima2002natural, lee2008save, khan2011human, khan2011towards, hanckmann2012automated, barbu2012video} focus on rule-based systems where sentences are generated using predefined templates following certain linguistic rules. Later, learning-based data driven approaches \citep{das2013thousand, guadarrama2013youtube2text, krishnamoorthy2013generating, sun2014semantic, rohrbach2014coherent, rohrbach2013translating, xu2015jointly, yu2015learning} became popular. As the methods started becoming free from manual engineering, the problem became more scalable providing flexibility to work with larger datasets. Recently, Recurrent Neural Network (RNN)-based approaches \citep{donahue2015long, venugopalan2015sequence, venugopalan2014translating, yao2015describing, xu2015multi} have achieved promising performance in video captioning. One of the earliest works \citep{venugopalan2014translating} using RNNs extends the image captioning methods by average pooling the video frames which only works for short video clips containing just one event. To overcome this shortcoming, recurrent encoder-based models \citep{donahue2015long}, \citep{xu2015multi}, \citep{venugopalan2015sequence}  and attention models \citep{yao2015describing} have been proposed. \citep{yu2016video} uses a hierarchical RNN to generate a paragraph for richer description. Another paper \citep{krishna2017dense} performs dense-captioning of events in videos using context information. There is only one other previous work \citep{sener2019zero} for predicting the captions of future unobserved activities. We want to point out a major difference between their work and ours. Their work uses the caption from the observed portion of the video to predict the caption for the future unobserved events. This setting can be restrictive for some real-world scenarios like autonomous driving, assistive systems for the visually impaired or human-robot interaction where only visual data captured by the camera is available. Our method is based on a two-step prediction approach where the action labels are predicted first using only the visual spatio-temporal context from the observed events and then the predicted labels and the observed scene context are used for caption prediction. We do not need any captions on the observed portion of the video. Thus, our approach tackles a more challenging and generic problem. 

{\bf Extension to Previous Works.}
The goal of this work is to predict the captions of a sequence of future activities which is, to the best of our knowledge, one of the earliest works in this area. This is leveraged on our previously published paper on activity label and starting time prediction \citep{mahmud2017joint}. Instead of predicting the label of one future activity at a time, here we are predicting the labels of a sequence of future activities and finally captioning the future activity sequence using the predicted label information. We show results on two additional datasets called MPII-Cooking $2$ Dataset \citep{rohrbach2016recognizing} and EPIC-KITCHENS Dataset \citep{Damen2018EPICKITCHENS} for label prediction and conduct experiments on a video description dataset called TACoS Multi-Level Corpus \citep{rohrbach2014coherent} built on MPII-Cooking $2$ \citep{rohrbach2016recognizing} demonstrating the effectiveness of our captioning method.

\section{Methodology}
In this section, we discuss the motivation behind the choice of our networks explaining the importance of different context attributes for the task, the network architectures in details, the training scheme and the way we obtained the final results in the test phase.

\begin{figure*}[t!]
	\centering
	\includegraphics[width=0.8\linewidth] {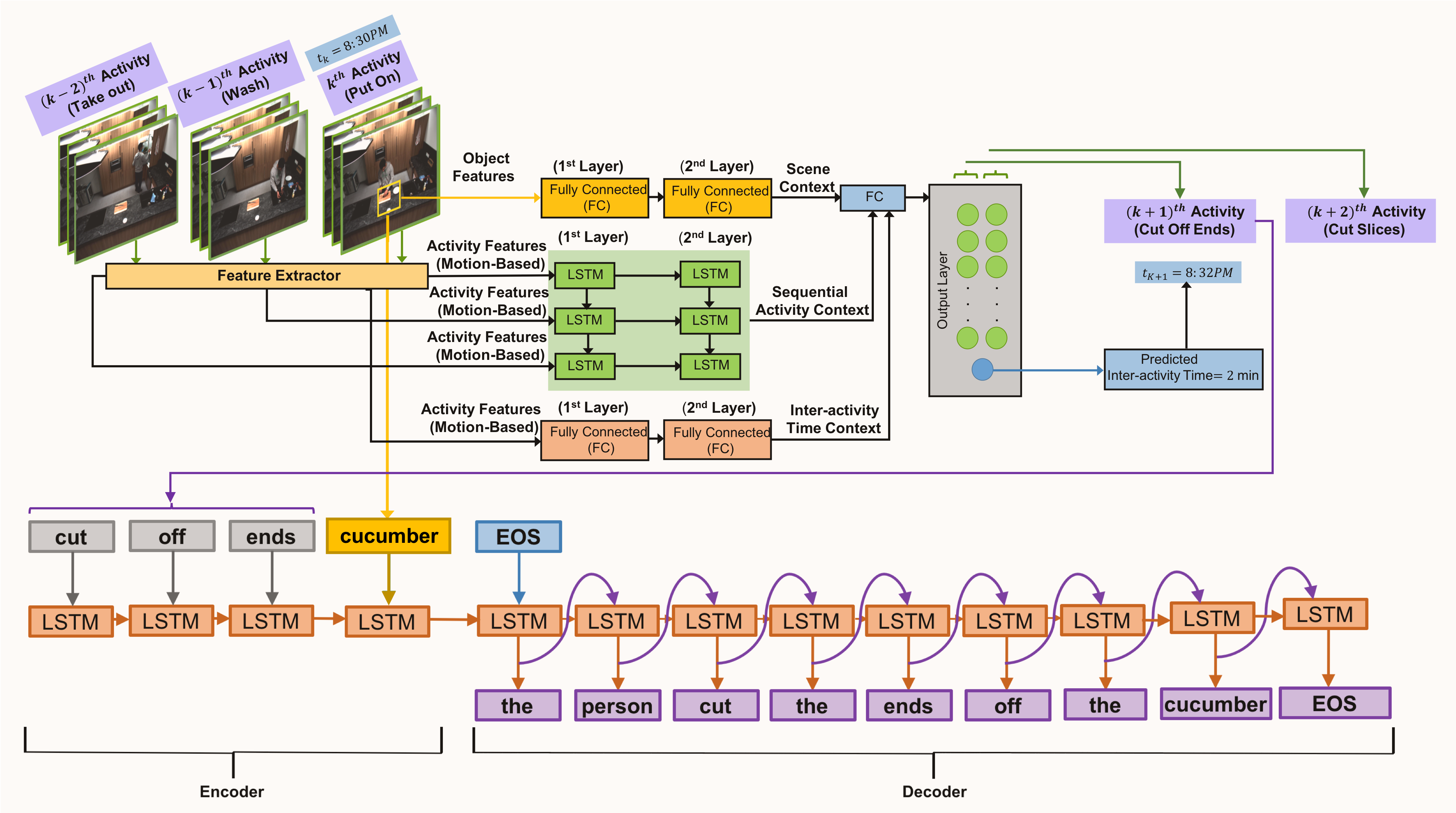}
	\caption{Proposed architecture for future activity label and caption prediction. In the top figure, the first two fully connected layers (yellow) incorporate the scene context (extracted from objects in the last frame) which use object features as input. The two LSTM layers (green) are used to incorporate the sequential activity context (`take out', `wash', `put on' in this example) which use motion-based features as inputs. The last two fully connected layers (peach) are used to incorporate inter-activity time context which use the last observed activity features (motion-based) (`put on' in this example) as input. There is a fully connected layer (blue) where all these layers are merged together. The output layer (gray) performs the final prediction, where for each element of the future activity sequence, the first few nodes (green) are used as the logistic regression nodes for label prediction. The last node (blue) of the output layer is used as the regression node for starting time prediction.  All of the layers have $256$ nodes. In the bottom figure, the predicted label (`cut off ends' in this example) and the scene context (`cucumber' in this example) are then used as input to the encoder LSTM layers and finally the decoder LSTM layers generate the captions. Here, EOS denotes End of Sentence.}
	\label{fig:fig4}
\end{figure*}

\subsection{Label Prediction for Activity Sequences} 
\subsubsection{{\bf Role of Different Context Attributes}} \label{section:Context1}
Activities usually have some temporal ordering, e.g., a person will get into a car and then it starts to move, or vegetables will be washed, peeled, and then put on the pan. Therefore, previous activities can provide useful information about the upcoming ones which can be referred to as {\bf sequential activity context}. It is true that not all the steps in a complex activity will always occur exactly in the same sequence, which is why our prediction results are probabilistic and different outputs are possible. Activities are also characterized by the objects present in the last observed portion of the scene during the time of their occurrence which can be referred to as {\bf scene context}.  For many activities, predicting the future has multiple plausible options. To reduce this specific ambiguity, we take scene context into account along with the sequential information. Thus, combining the information obtained from these two different context attributes (temporal sequence and spatial objects), we infer the sequence of future unobserved activities. For example, if three sequential activities in a video are `wash objects', `peel' and `cut slices', then there may be two probable future activity sequences: `screw open', `take out from spice holder', and `spice' or `put in bowl', `puree' and `smell' (based on two different training instances). But a bowl present in the scene increases the possibility of the latter sequence. 

Several research works on activity recognition \citep{choi2011learning, deng2015deep, ibrahim2015hierarchical, lan2010beyond, wang2015video, yao2010modeling, zhu2013context} and prediction \citep{chakraborty2014context, abu2018will} have shown significant performance improvement by using such context information which are also known as context-aware approaches. Most of the existing works have graphical model-based approaches for context incorporation. However, they are not very suitable to handle the context of long-term dependency. As mentioned before, LSTM is a popular choice for sequential context incorporation. LSTM networks are straightforward to fine-tune end-to-end and can handle sequential data of varying lengths. So, we use LSTM to incorporate sequential activity context. However, for including the scene context from the last observed frame, there is no need for handling such sequential dependency (taking long-term scene context into account can increase ambiguity) and fully connected layers can capture this efficiently.

The inter-activity time between different activities depends on their labels. For example, we know from  experience that `peel' or `cut slices' takes more time than `wash objects'. Thus, by observing the previous activity features we can infer the inter-activity time (difference between the starting time of the observed activity and the future activity) which can be referred to as {\bf inter-activity time context}. 


\subsubsection{{\bf Network Architecture}} \label{section:Architecture1}
Our proposed architecture for jointly predicting the labels and the starting time of a future activity sequence is shown in Fig.~\ref{fig:fig4}. The LSTM is used to solve a sequential input, sequential output problem. We use the activity features extracted from three (chosen empirically) previously observed activities as the LSTM input. Increasing the sequence length does not improve the prediction accuracy significantly (see Section \ref{section:ObservationHorizon} for details). We use a two-layer (chosen empirically) LSTM with $256$ memory units in each layer. The input of the two (chosen empirically) fully connected layers in the first branch are the visual features extracted from the objects present in the last observed portion of the scene and there are $256$ nodes in each layer. The input of the two (chosen empirically) fully connected layers in the third branch are the activity features extracted from the last observed activity and have $256$ nodes in each layer too. Parameter sensitivity analysis for the number of nodes used are provided in the supplementary material.

Finally, the outputs from these three branches are combined together and another fully connected layer is added on top of it. The merging combines the effect of different context attributes. In the output layer, for each of the three future activities in the sequence, the first few (equal to the number of activity classes) nodes are used as the logistic regression nodes for sequential label prediction and the last node is used as a regression node for predicting the starting time of the future activity sequence.

\subsubsection{{\bf Model Training Approach}}\label{section:Training1}
This training method differs from our previous approach \citep{mahmud2017joint} in terms of the training procedure. We use the popular open-source deep learning package Keras \citep{chollet2015keras} with TensorFlow \citep{abadi2016tensorflow} in the backend which has ready-to-use implementations of LSTM and fully connected layers.  The input sequences for the LSTM are chosen in a sliding window manner with a stride of one for data augmentation. For example, to predict the labels of the future sequence containing the $(k+1)^{th}$, $(k+2)^{th}$ and $(k+3)^{th}$ activities, activity features extracted from the $k^{th}$, $(k-1)^{th}$ and $(k-2)^{th}$ activities are used. For predicting the labels of the future sequence containing the $(k+2)^{th}$, $(k+3)^{th}$ and $(k+4)^{th}$ activities, activity features extracted from the $(k+1)^{th}$, $k^{th}$ and $(k-1)^{th}$ activities are used and so on. We have not used the ground-truth labels of the observations during either training or testing for any of the experiments. The two fully connected layers in the first branch use visual object features from the last observed portion of the scene as input. Another two fully connected layers in the third branch use activity features extracted from the last observed activity as input. We use ReLU activation function for all the fully connected layers. In the output layer, we use softmax activation function in the logistic regression nodes for predicting the label of each activity in the sequence and ReLU activation function in the regression node for predicting the starting time of the sequence. The parameters of the entire network are jointly optimized. 


We take the summation of the following two losses to compute the final loss. One is the cross-entropy loss function which is defined as follows:
\begin{eqnarray}
\mathcal{L}(\mathbf{X}, \mathbf{Y})  &= -\frac{1}{n}\sum_{i = 1}^{n}\sum_{j= 1}^{c} \mathbf{1}(y^{(i)} = j) \nonumber \\ 
&\times \log p(y^{(i)} = j|\mathbf{x}^{(i)})
\end{eqnarray}
Here, $\mathbf{X} = \{\mathbf{x}^{(1)}, ... , \mathbf{x}^{(n)}\}$ is the set of input feature vectors (activity features of the last three observed activities and features of the objects present in the last observed portion of the scene) in the training dataset, $\mathbf{Y} = \{y^{(1)}, ... , y^{(n)}\}$ is the corresponding set of labels for those input features, and $j = \{1, ..., c\}$ is the set of class labels. $\mathbf{1(.)}$ is an indicator function. $\mathbf{x}^{(i)}$ is the sequential activity features extracted from the previous three activities.

Another is the mean squared loss between the ground truth inter-activity time and the predicted inter-activity time which is defined as follows:
\setlength{\abovedisplayskip}{8pt}
\setlength{\belowdisplayskip}{8pt}
\begin{eqnarray}
\mathcal{L}(\mathbf{P}, \mathbf{Q})  = \frac{1}{n}\sum_{i = 1}^{n} (q^{(i)}-\hat{q}^{(i)})^2
\end{eqnarray}
Here, $\mathbf{P} = \{\mathbf{p}^{(1)}, ... , \mathbf{p}^{(n)}\}$ is the set of input feature vectors (activity feature of the last observed activity) in the training dataset, and $\mathbf{Q} = \{q^{(1)}, ... , q^{(n)}\}$ is the corresponding set of inter-activity times for those input features. $\hat{q}^{(i)}$ represents the predicted inter-activity time given input $p^{(i)}$ where the ground truth inter-activity time is $q^{(i)}$. $\hat{q}^{(i)}$ is a function of the input features $\mathbf{P}$. The outputs of the training are the labels of the three future activities and the starting time of that activity sequence.

The parameters of the network are jointly optimized by minimizing both of these losses. To optimize the network, we use a stochastic gradient descent with an adaptive sub-gradient method (Adam) \citep{kingma2014adam} which is popular for its impressive history of empirical success. We also tested with Adagrad \citep{duchi2011adaptive}, Adamax \citep{kingma2014adam}, Nadam \citep{dozat2016incorporating} and RMSProp \citep{tieleman2012lecture} but empirically chose Adam. We use Dropout layer \citep{srivastava2014dropout} with a probability of $0.2$ after each layer to prevent overfitting. The batch size is set to $128$. We use a learning rate of $0.001$.

\subsection{Caption Generation for Activity Sequences}  

\subsubsection{{\bf Role of Scene Context for Label to Caption Mapping}} \label{section:Context2}
Motivated by the inspiring performance of the sequence-to-sequence models in \citep{sutskever2014sequence} for machine translation and in \citep{venugopalan2015sequence} for video to text mapping, we use a similar model for label to sentence mapping where both the input ${(\mathbf{a_1}, \mathbf{a_2}, \cdots, \mathbf{a_m})}$ and the output ${(\mathbf{b_1}, \mathbf{b_2}, \cdots, \mathbf{b_n})}$ are sequences of words of variable length. Since the labels do not contain any object information, it is hard to predict the object in the caption only from the label. For example, it is difficult to map from \emph{wash} to \emph{A person washed carrots}. So, we use the scene context from the observed portion along with the label in the encoder LSTM input for meaningful mapping of the objects.

\subsubsection{{\bf Network Architecture}} \label{section:Architecture2}
The input to the encoder LSTM is text e.g., \textit{cut apart cucumber}, \textit{take out egg fridge}, \textit{cut off ends carrot}, etc. obtained by concatenating the corresponding predicted labels and scene context. In the captions, verbs are followed by objects. To maintain this order, scene context follows the label in the text input. So, sequence-to-sequence learning via encoder LSTM is important here to incorporate this sequential information efficiently and maintain meaningful structure between subject, verb and objects. We do not provide subject as the text input since the subject is constant (\textit{the person}) throughout the dataset. However, for any other dataset where different subjects exist e.g., \textit{man}, \textit{woman}, \textit{boy}, \textit{girl} etc., our network would take the text input in subject-verb-object order as a natural structure. An encoder-decoder LSTM pair is the best option for maintaining meaningful structure between subject, verb and object to incorporate this information correctly. Both the encoder LSTM and the decoder LSTM have $3$ layers with $1000$ memory units in each layer. 

\subsubsection{{\bf Model Training Approach}} \label{section:Training2}
At first, we perform embedding by generating a dictionary using all the words in the input of the training set and then convert these words to one hot vectors according to that dictionary. We choose a vocabulary size of $20000$. We use one LSTM layer to encode the label and the scene context to a fixed-dimensional vector and use another LSTM layer to generate a sentence from that vector. We estimate the conditional probability ${p(\mathbf{b_1}, \mathbf{b_2}, \cdots, \mathbf{b_n}}|{\mathbf{a_1}, \mathbf{a_2}, \cdots, \mathbf{a_m})}$ given the input ${(\mathbf{a_1}, \mathbf{a_2}, \cdots, \mathbf{a_m})}$. In our case, since the caption is always longer than the combination of label and scene context, $n$ is always bigger than $m$.

During encoding, the first LSTM generates a sequence of hidden states ${(\mathbf{h_1}, \mathbf{h_2},\cdots, \mathbf{h_m})}$ given the label and the scene context ${(\mathbf{a_1}, \mathbf{a_2}, \cdots, \mathbf{a_m})}$. Then a fixed-dimensional vector $\bf{z}$ corresponding to the label is generated by the last hidden state of the LSTM. The decoder LSTM computes the conditional probability of the output sentence given the input label and the scene context as follows:
\begin{align}
&p\left(\mathbf{b_1}, \mathbf{b_2}, \cdots, \mathbf{b_n}|\mathbf{a_1}, \mathbf{a_2}, \cdots, \mathbf{a_m}\right) \nonumber \\
&~~~~~~~~~~~~~~~~~~~~~~~~~~~~~~~~~~~~=\prod_{d = 1}^{n}p\left(\mathbf{b_d}|\mathbf{z}, \mathbf{b_1}, \cdots, \mathbf{b_{d-1}}\right)
\end{align}
where the distribution of $p(\mathbf{b_d}|\mathbf{z}, \mathbf{b_1}, \cdots, \mathbf{b_{d-1}})$ is represented by a softmax over all the words in the vocabulary.

During training, the log probability of a correct caption (sentence) is maximized given the label and the scene context. Cross-entropy loss function is used in this model. The batch size we use is $1000$. Keras \citep{chollet2015keras} with TensorFlow \citep{abadi2016tensorflow} is the library we use for this work. 

\subsection{Test Case Scenario}\label{section:Test} 
For predicting the labels of a future activity sequence, the activity features of the last three observed activities are used in the LSTM input, the features of the objects present in the last observed portion of the scene are used as the input of the first fully connected layers and the activity features of the last observed activity are used as the input of another two fully connected layer. Based on the learned sequence-to-sequence relationship in the training phase, the network predicts the labels of the next three unobserved activities. Using this predicted sequence of labels and observed scene context, the most likely captions for the future activity sequence are generated by the encoder-decoder LSTM pair. 

\section{Experiments}
\label{section:Experiments}
We conduct experiments on four challenging datasets: MPII-Cooking Dataset \citep{rohrbach2012database}, MPII-Cooking $2$ Dataset \citep{rohrbach2016recognizing}, EPIC-KITCHENS Dataset \citep{Damen2018EPICKITCHENS} (fine grained indoor activities) and VIRAT Ground Dataset \citep{oh2011large} (coarse outdoor activities) to evaluate the performance of our label prediction framework for a future activity sequence. We provide the starting time prediction performance for a future sequence from \citep{mahmud2017joint} in the supplementary material. To evaluate the performance of our proposed captioning framework, we conduct experiments on the challenging video description dataset TACoS Multi-Level Corpus \citep{rohrbach2014coherent} built on MPII-Cooking $2$ \citep{rohrbach2016recognizing}. The goal of the experiments is to compare our predictions with ground truth values as well as the state-of-the-arts, and perform an ablation analysis of the methods.

\subsection{Datasets}
{\bf MPII-Cooking Dataset.} MPII-Cooking Dataset is a fine grained complex activity dataset where the participants interact with different tools, ingredients, and containers to complete a recipe. It has $65$ different cooking activities recorded from $12$ participants. There are $44$ videos with a length of more than $8$ hours. The dataset contains a total of $5,609$ annotations \citep{rohrbach2012database}. 

{\bf MPII-Cooking $2$ Dataset.} MPII-Cooking $2$ Dataset is a fine grained complex activity dataset where the participants interact with different tools, ingredients, and containers to complete a recipe. It has $67$ different cooking activities recorded from $30$ participants. In total there are $273$ videos with a length of more than $27$ hours \citep{rohrbach2016recognizing}. 

{\bf EPIC-KITCHENS Dataset.} EPIC-KITCHENS Dataset is a large-scale egocentric video benchmark dataset comprised of cooking activities performed by $32$ different subjects in their native kitchen environments. The dataset contains $55$ hours of video consisting of $11.5M$ frames, $272$ training videos, $39596$ annotated segments, and $454.3K$ object bounding boxes. Considering  all unique pairs of verb and noun in the public training set, there are total
 $2513$ unique actions \citep{Damen2018EPICKITCHENS}. 

{\bf VIRAT Ground Dataset.} VIRAT Ground Dataset is a challenging human activity dataset which consists of $11$ different activities recorded in natural outdoor scenes with background clutter. There are total $329$ videos with a length of around $5$ hours \citep{oh2011large}. However, we use only $275$ of them as some videos have incomplete annotations. 

{\bf TACoS Multi-Level Corpus.} This video description dataset consists of $185$ long indoor videos which contain different actors, fine-grained activities, and small objects in daily cooking scenarios. Each video is annotated by multiple turkers. For each video, there are detailed descriptions with at most $15$ sentences, a short description ($3$-$5$ sentences), and a single sentence. Since, workers could describe videos without aligning each sentence to the video, the descriptions are natural and have a complex sentence structure \citep{rohrbach2014coherent}. 
\begin{table*}[t!]
	\caption{Label recognition performance for all of the datasets. The best numerical results are marked in bold.}
	\begin{center}
		\begin{tabular}{|M{3.20cm}|M{3.80cm}|M{1.3cm}|M{1.0cm}|M{1.8cm}|M{1.8cm}|M{1.8cm}|}
			\hline
			\textbf{Dataset}   & \textbf{Method}  & \textbf{Precision} & \textbf{Recall}    & \textbf{Accuracy \% (Top-1)} & \textbf{Accuracy \% (Top-3)} & \textbf{Accuracy \% (Top-5)}\\
			\hline \hline
			\multirow{2}{*}{MPII-Cooking \citep{rohrbach2012database}} & CNN + LSTM \citep{ni2016progressively}  & 34.8  & 51.7  & -    & -    & -  \\
	         & 	\textbf{Proposed Method}  & \textbf{72.1}  & \textbf{69.3}  & \textbf{81.3} & \textbf{93.3}                                 & 	\textbf{96.2}  \\
			\hline    \hline     
				\multirow{2}{*}{MPII-Cooking $2$ \citep{rohrbach2016recognizing}}  & Dense trajectories \citep{rohrbach2016recognizing}   & 52.2  & -   & -   & -   & -  \\
			 & \textbf{Proposed Method} & \textbf{61.6}  & \textbf{55.6}    & \textbf{69.0} &\textbf{88.3}  &\textbf{92.7} \\
        	\hline \hline
        	\multirow{2}{*}{EPIC-KITCHENS \citep{Damen2018EPICKITCHENS}} & Fusion \citep{miech2019leveraging}  & -  & -  & 6.0 & - & -  \\
        	& 	\textbf{Proposed Method}  & \textbf{3.9}  & \textbf{3.5}  & \textbf{11.4} & \textbf{21.4} & \textbf{27.2}  \\
        		\hline    \hline 
				\multirow{2}{*}{VIRAT  \citep{oh2011large}}  & Sparse Autoencoder \citep{hasan2014continuous}   & -   & -  & 54.2    & -  & -   \\
		     & \textbf{Proposed Method} & \textbf{52.1}  & \textbf{25.2}  & \textbf{71.9}  & \textbf{86.5}  & \textbf{95.4} \\
			\hline                                                   
		\end{tabular}
	\end{center}
	\label{tab:Table1}
\end{table*}
\begin{table*}[t!]
	\caption{Label prediction performance for all of the datasets. The best numerical results are marked in bold.}
	\begin{center}
	\begin{tabular}{|M{3.20cm}|M{3.80cm}|M{1.3cm}|M{1.0cm}|M{1.8cm}|M{1.8cm}|M{1.8cm}|}
		\hline
		\textbf{Dataset}   & \textbf{Method}  & \textbf{Precision} & \textbf{Recall}    & \textbf{Accuracy \% (Top-1)} & \textbf{Accuracy \% (Top-3)} & \textbf{Accuracy \% (Top-5)}\\
		\hline \hline
			\multirow{2}{*}{MPII-Cooking \citep{rohrbach2012database}} & Convolutional LSTM \citep{quteprints118022} & -   & -  & 60.4 & -  & - \\ 
		& 	\textbf{Proposed Method}  & \textbf{72.1}  & \textbf{67.6}  & \textbf{79.9} & \textbf{93.0}                                 & 	\textbf{96.2}  \\
		\hline    \hline     
			\multirow{2}{*}{MPII-Cooking $2$ \citep{rohrbach2016recognizing}}  & -   & -  & -   & -   & -   & -  \\
		& \textbf{Proposed Method} & \textbf{58.8}  & \textbf{53.3}    & \textbf{65.5} &\textbf{82.4}  &\textbf{88.5} \\
		\hline \hline
			\multirow{2}{*}{EPIC-KITCHENS \citep{Damen2018EPICKITCHENS}} & \textbf{RU-LSTM \citep{furnari2020rolling}}  & -  & -  & - & - & \textbf{35.3}  \\
		& 	Proposed Method  & 6.6  & 3.7 & 11.3 & 23.6 & 29.6 \\
		\hline    \hline 
			\multirow{3}{*}{VIRAT  \citep{oh2011large}}  & Graphical Model \citep{chakraborty2014context}  & -   & -  & 68.5    & -  & -   \\
		& SCR-Graph \citep{chen2019scr} & 21.3   & -  & -  & -  & - \\
		& \textbf{Proposed Method} & \textbf{49.6}  & \textbf{22.2}  & \textbf{71.8}  & \textbf{86.4}  & \textbf{94.4} \\
		\hline                                                   
	\end{tabular}
\end{center}
	\label{tab:Table2z}
\end{table*}
\begin{table*}[h!]
	\caption{Sequence prediction performance comparisons for all of the datasets. The best numerical results are marked in bold.}
	\begin{center}
		\begin{tabular}{|M{3.20cm}|M{3.50cm}|M{3.5cm}|M{4.5cm}|}
			\hline
			\textbf{Dataset}   & \textbf{Method}  & \textbf{Accuracy \% Next-to-Next Activity} & \textbf{Accuracy \% Next-to-Next-to-Next Activity} \\
			\hline \hline
				\multirow{2}{*}{MPII-Cooking \citep{rohrbach2012database}} & Multi-step Prediction & 78.1    & 77.5  \\
			& 	\textbf{Proposed Method}  & \textbf{79.1}  & \textbf{78.1}  \\
			\hline    \hline     
				\multirow{2}{*}{MPII-Cooking $2$ \citep{rohrbach2016recognizing}}  & Multi-step Prediction  & 63.7   & 62.6  \\
			& \textbf{Proposed Method} & \textbf{64.4}  & \textbf{63.5}  \\
			\hline \hline
				\multirow{1}{*}{EPIC-KITCHENS \citep{Damen2018EPICKITCHENS}}  
			& \textbf{Proposed Method} & \textbf{8.1}  & \textbf{6.6}  \\
			\hline \hline
				\multirow{2}{*}{VIRAT  \citep{oh2011large}}  & Multi-step Prediction  & 70.7  & 68.5   \\
			& \textbf{Proposed Method} & \textbf{71.5}  & \textbf{69.2} \\
			\hline                                                   
		\end{tabular}
	\end{center}
	\label{tab:Table2}
\end{table*}

Detailed description of these datasets are available in the supplementary material. These datasets are untrimmed, unlike the trimmed datasets popularly used for recognition tasks in activity analysis, and have context information. Since we are captioning unobserved future activities, we need untrimmed datasets containing natural sequences of activities with annotated video description. Because of these requirements, the choice of datasets on which our method can be demonstrated is limited. For example, we cannot use MPII-Cooking Dataset \citep{rohrbach2012database} or VIRAT Ground Dataset \citep{oh2011large} for captioning evaluation as used in \citep{mahmud2017joint} for label prediction evaluation since they do not have human descriptions. We cannot use YouCookII Dataset \citep{zhou2017procnets} or Tasty Dataset \citep{sener2019zero} since they do not have the action label annotation required for training the label prediction network. We cannot use Activity Net Captions \citep{krishna2017dense} because there are only $1.5$ activity instances on average in each video which is not enough to incorporate the sequential context for label prediction.

\subsection{Features}
We use C3D features pre-trained on the Sports-1M dataset \citep{tran2015learning} as activity features for MPII-Cooking Dataset, MPII-Cooking $2$ Dataset, and VIRAT dataset. The C3D features are of size $4096$ and extracted for each $16$ frames with a temporal stride of $8$ frames. Then we perform max pooling to get a fixed-length feature vector for the video. For EPIC-KITCHENS, we use the pre-extracted RGB and flow features for the sequential activity context and the faster R-CNN-based object features for the scene context provided by \citep{furnari2019would,furnari2020rolling}. We claim that our method is independent of any particular choice of feature. This is mentioned in Section \ref{section:Labelresult} where using bag-of-word based Motion Boundary Histograms (MBH) \citep{dalal2006human} features gives similar label prediction results for MPII-Cooking Dataset. According to \citep{wang2011action}, MBH features are extracted around densely sampled points and a codebook is generated using k-means clustering for these $4000$ words long features. 



\subsection{Label Prediction Results for Activity Sequences} \label{section:Labelresult}

{\bf Objective.}
The main objective of these experiments is to analyze how well our framework can predict the labels of a future unobserved activity sequence. 

{\bf Performance Measures.}
The evaluation metrics we use are:  \begin{enumerate*} \item multi-class precision (Pr), \item multi-class recall (Rc), and \item overall accuracy \end{enumerate*} for top-$1$ match, top-$3$ matches and top-$5$ matches. For all these metrics, the higher value indicates better prediction performance. 

{\bf Compared Methods.}
We compare with existing recognition approaches by using our method to determine the labels of the observed activities. For recognition of the observed activities, we use the observed activity features from the $(i-2)^{th}$, $(i-1)^{th}$ and $i^{th}$ activities to predict the label of the $i^{th}$ activity and so on. For MPII-Cooking Dataset, we compare with \citep{ni2016progressively} which estimates the labels of the observed activities. We show that the performance of our prediction of observed activities, is better than that of the recognition method using a combination of CNN and LSTM \citep{ni2016progressively}. For label prediction, we compare with a convolutional LSTM-based approach \citep{quteprints118022} and achieve better performance. For MPII-Cooking 2 Dataset, we compare with \citep{rohrbach2016recognizing} which estimates the labels of the observed activities and show that the precision we achieve for prediction of the observed activities, is higher than that of the recognition method using a combination of dense trajectories and hand trajectories \citep{rohrbach2016recognizing}. For EPIC-KITCHENS Dataset, we compare with \citep{miech2019leveraging} which estimates the labels of the observed activities and show that the accuracy we achieve for prediction of the observed activities, is higher than that of the recognition method using a model constrained to reason about the present \citep{miech2019leveraging}. For label prediction, we compare with the state-of-the-art rolling-unrolling LSTM-based approach \citep{furnari2020rolling} and achieve comparable performance. Please note that we cannot compare recognition performance with \citep{furnari2020rolling} since they reported results only on the test set of EPIC-KITCHENS Dataset for which the ground truth is not available since the test evaluation server is closed now. For VIRAT Dataset, there is a graphical model-based approach \citep{chakraborty2014context}, a semantic scene labeling-based approach \citep{kitani2012activity}, and a graph reasoning network-based approach \citep{chen2019scr} for prediction of future unobserved activities. We compare our method with \citep{chakraborty2014context} and \citep{chen2019scr} and achieve higher accuracy for label prediction. We cannot compare with \citep{kitani2012activity} since they use scene specific customized set of labels which are not annotated in the original dataset. We also compare with a state-of-the-art active learning-based recognition approach which uses sparse autoencoder \citep{hasan2014continuous} and achieve higher accuracy.

To evaluate our label prediction results for further activities in the future sequence, we compare with our previous multi-step prediction baseline \citep{mahmud2017joint} where we predicted the next-to-next activity i.e., $2$-step prediction (using activity features from  the $(i-3)^{th}$, $(i-2)^{th}$ and $(i-1)^{th}$ activities, we predicted the label of the $(i+1)^{th}$ activity) and the next-to-next-to-next activity i.e., $3$-step prediction. Multi-step prediction is different from sequence prediction. In multi-step prediction, each prediction step is treated as uncorrelated with the others, while in sequence prediction, the correlations are accounted for.  For example, in multi-step prediction, we predict the label of the $(i+1)^{th}$ unobserved activity using context information from the $i^{th}$, $(i-1)^{th}$, and $(i-2)^{th}$ activities one step at a time. But in sequence prediction, we predict the labels of the $i^{th}$, $(i+1)^{th}$, and $(i+2)^{th}$ activities together using context information from the $(i-1)^{th}$, $(i-2)^{th}$, and $(i-3)^{th}$ activities. Since these individual activities are part of a single sequence and thus correlated with each other, the network learns the temporal ordering of the activities better when a sequence of activities is used as the output during training. We also compare our sequence prediction results with \citep{abu2018will}.

{\bf Experimental Setup.}
For experiments on MPII-Cooking Dataset, we use five fold leave-one-person-out cross validation approach. Among $12$ subjects, we use $7$ for training and $5$ for testing. For each of the five training instances, we use $7$ training subjects and $4$ testing subjects for training, leaving $1$ from that set for testing. This is done $5$ times leaving $1$ testing subject out and then averaging the results known as ``five-fold leave-one-person-out" cross validation. For experiments on MPII-Cooking $2$ Dataset, we use the experimental setup (same train-test split) of \citep{rohrbach2016recognizing}. For activity recognition experiments on EPIC-KITCHENS Dataset, we use the experimental setup (same train-test split) of \citep{miech2019leveraging} i.e. report results on the same validation set composed of the following kitchens: P03, P14,
P23, and P30 from the public training set since the labels of the test sets (S1 and S2) are not
available in the evaluation server anymore. For activity prediction experiments on EPIC-KITCHENS Dataset, we use the experimental setup ($232$ training videos and $40$ validation videos) and protocol ($1$ sec anticipation time) of \citep{furnari2020rolling}. For experiments on VIRAT Ground Dataset, we use the first $170$ videos for training and the rest of them for testing. The network is trained on a NVIDIA Tesla K80 GPU. 

{\bf Recognition Performance.} 
Comparisons of our recognition performance (for observed activities) on all of the datasets with the state-of-the-art recognition methods are shown in Table~\ref{tab:Table1}. The methods we compare to did not report all of the evaluation metrics we use - hence the missing values. Our method outperforms the state-of-the-arts for all of the datasets.

{\bf Prediction Performance.} 
Our label prediction results (for unobserved future activities) for all of the datasets and comparison with other prediction methods for MPII-Cooking Dataset, EPIC-KITCHENS Dataset, and VIRAT Ground Dataset are shown in Table~\ref{tab:Table2z}. It is seen that our method outperforms the prediction method proposed in \citep{quteprints118022} for MPII-Cooking, achieves comparable prediction performance with \citep{furnari2020rolling} for EPIC-KITCHENS Dataset, and outperforms the prediction method proposed in \citep{chakraborty2014context} and \citep{chen2019scr} for VIRAT. The methods we compare to did not report all of the evaluation metrics we use - hence the missing values. We achieve similar label prediction accuracy of $79.9\%$ and $80.7\%$ for MPII-Cooking Dataset using C3D and MBH features respectively which justifies the claim that our method is independent of the choice of features. Sequence prediction result comparisons with the baseline multi-step prediction method are shown in Table~\ref{tab:Table2}. For EPIC-KITCHENS Dataset, there might be other actions which are not annotated between two annotated actions. Since the dataset is not densely annotated, it is not meaningful to use the dataset for multi-step prediction because of the lack of reliable ground truth. Therefore, we do not report multi-step prediction results for EPIC-KITCHENS Dataset. As the prediction horizon increases, there is a gradual accumulation of error. It is to be noted that even for sequence prediction, prediction results for the first activity in the future sequence have higher accuracy than that of the next activities as we are still using \textbf{scene context} from the last observed portion of the scene which is related to the immediate future activity label. Using the RNN-based anticipation approach of \citep{abu2018will} with ground truth observations, we achieve a Top-$1$ accuracy of only $26.3\%$ for MPII-Cooking Dataset. This is expected since \citep{abu2018will} does not incorporate scene context or inter-activity context and the effect of these context attributes are shown in Table~\ref{tab:Table5} in Section \ref{section:Context3}.



\subsubsection{\bf Multiple Possibilities for Future Activity Labels}
One particular activity sequence can have multiple possible outcomes. For example, `wash objects' and `peel' can be followed by either `cut apart' and `cut slices'. As the network has been trained on both of these possible sequences (in one case the network has probably seen `cut apart' as the next activity and in another case `cut slices' as the next activity), it is hard to say precisely which is the next activity. Earlier we mentioned that in case of multiple possibilities, such as while choosing between `spice' or `put in bowl' after `wash objects', `peel' and `cut slices', a bowl in the scene increases the probability of the activity label being the latter one. But in these types of closely related activities (`cut apart' and `cut slices'), scene context cannot contribute much as both of the activities require a knife. Therefore, we present the top-$k$ choices with the associated probabilities for each of them. In spite of having many closely related ambiguous activities (`cut dice', `cut slices', `cut apart') in the dataset, our top-$1$ match outperforms the baseline in terms of accuracy. Our method can also handle an `unknown' label (never seen in training) when the probability of none of the predicted future activities crosses a threshold.
\subsubsection{\bf Effect of Different Context Attributes.}\label{section:Context3}
We perform an ablation study to justify the choice of our network. Using only sequential activity context and scene context (eliminating inter-activity time context), we get relatively lower label prediction accuracy for all of the datasets than that of our proposed network. Similarly, using only sequential activity context and inter-activity time context (eliminating scene context), we get lower label prediction accuracy than that of our proposed network for all of the datasets as shown in shown in Table~\ref{tab:Table5}. 
\begin{table}[h]
	\caption{Ablation study for label prediction for all of the datasets. The best numerical results are marked in bold.}
	\begin{center}
		\begin{tabular}{|M{1.6cm}|M{1.4cm}|M{1.9cm}|M{1.9cm}|}
			\hline
			\multirow{4}{*}{\textbf{Dataset}} & \multicolumn{3}{c|}{\textbf{Top-1 Accuracy\%}}    \\ \cline{2-4}
			& \multirow{3}{*}{\begin{tabular}[c] {@{}c@{}}\textbf{Proposed}\\ \textbf{Network}\end{tabular}} & \multirow{3}{*}{\begin{tabular}[c] {@{}c@{}} \textbf{Removing} \\ \textbf{inter-activity} \\ \textbf{time context} \end{tabular}} & \multirow{3}{*}{\begin{tabular}[c] {@{}c@{}} \textbf{Removing} \\ \textbf{scene context} \end{tabular}} \\
			& & & \\
			& & & \\ \hline \hline
			{MPII-Cooking}   & \textbf{79.9}             & {75.7}                                 & {33.7}                   \\\hline
			{MPII-Cooking 2} & \textbf{65.5}             & {60.2}                                 & {45.7}                  \\ \hline
			{EPIC-KITCHENS} & \textbf{11.3}             & {8.7}                                 & {4.1}                  \\ \hline
			{VIRAT}   & \textbf{71.8}             & {69.2}     & {61.0}                   \\\hline                             
		\end{tabular}
	\end{center}
	\label{tab:Table5}
\end{table}

\subsubsection{\bf Analysis of Observation Horizon}\label{section:ObservationHorizon}
Here, we justify the choice of the observation horizon. We empirically chose a sequence length of $3$ for preceding activity features as sequence length of $2$, $5$, $7$ and $9$ give relatively lower accuracy for MPII-Cooking Dataset as shown in Table~\ref{tab:Table3}. We believe that this is because the longer horizon brings in the effect of activities that are not closely related.

\begin{table}[h]
	\caption{Sequence length sensitivity analysis for MPII-Cooking Dataset. The best numerical result is marked in bold.}
	\vspace{-2mm}
	\begin{center}
		\resizebox{\columnwidth}{!}{%
			\begin{tabular}{|M{1.2cm}|M{1.2cm}|M{1.2cm}|M{1.2cm}|M{1.2cm}|}
				\hline 
				\multicolumn{5}{|c|}{\textbf{Top-$1$ Accuracy \%}}                                                                                                                                                                                                                                                                                                    \\ \hline \hline
				\multicolumn{5}{|c|}{\textbf{Sequence Length}}                                                                                                                                \\ \hline
			\textbf{2}                                                         & \textbf{3}                                                          & \textbf{5}                                                           & \textbf{7}                                                            & \textbf{9} \\ \hline
				78.6                                                          & \textbf{79.9}                                                           & 79.1                                                            & 77.5                                                            & 76.9      \\ \hline
			\end{tabular}
		}
	\end{center}
	\label{tab:Table3}
\end{table}

\subsection{Captioning Results for Activity Sequences} \label{section:Captionresult}

{\bf Objective.}
The objective of these experiments is to evaluate the quality of our generated captions against the ground truth captions annotated by the human annotators. More results are presented in the supplementary material.

{\bf Performance Measure.}
The evaluation metrics we use are BLEU (Bilingual Evaluation Understudy) \citep{papineni2002bleu}, CIDEr (Consensus-based Image Description Evaluation) \citep{vedantam2015cider}, METEOR (Metric for Evaluation of Translation with Explicit ORdering) \citep{banerjee2005meteor}, ROUGE (Recall-Oriented Understudy for Gisting Evaluation) \citep{lin2005recall}, and SPICE (Semantic Propositional Image Caption Evaluation) \citep{anderson2016spice}. BLEU is a weighted average of variable length phrase matches against the reference translations in machine translation. CIDEr evaluates how well a candidate sentence matches the consensus of a set of image descriptions. METEOR uses the generalized concept of unigram matching between the machine produced translation and human-produced reference translations. ROUGE compares an automatically produced caption against a human-produced reference. We use ROUGE-L which naturally considers sentence level structure similarity and automatically identifies longest co-occurring in sequence n-grams. SPICE is an automatic image caption evaluation metric which compares semantic propositional content. In our case, the number of word matches is compared between the generated captions and the reference captions annotated by the human annotators. For all of the metrics, higher value indicates better performance. 

{\bf Comparisons.}
To the best of our knowledge, no existing method for generating captions for future unobserved events in videos has reported performance on TACoS Multi-Level Corpus. Therefore, as a baseline, we compare with the input text of the encoder LSTM with the ground truth captions in order to measure the improvement achieved by our caption generation module. We also compare with existing captioning approaches for the observed portion of the video by using our method to determine the captions of the observed activities. We compare with \citep{rohrbach2014coherent} which first predicts a semantic representation (SR) of the \emph{observed} portion and then generates detailed captions. We compare against their per sentence BLEU@$4$ score for short descriptions. We also compare with the BLEU@$4$, CIDEr and METEOR scores reported in \citep{yu2016video} which exploits hierarchical RNNs to generate captions for the observed portion.

Similar to label prediction, since none of the existing methods perform sequence prediction for captions, we can only compare our captioning result for the first unobserved activity with different state-of-the-art methods. However, to evaluate our captioning results for further activities in the future sequence, we compare with multi-step prediction baseline where we predict the next-to-next caption i.e., $2$-step caption prediction and the next-to-next-to-next caption i.e., $3$-step caption prediction. Multi-step captioning yields different results than sequence captioning because of the same reason as in label prediction. 

%
\begin{table}[t]
	\caption{Comparisons of BLEU@$4$ (in percent) (B), CIDEr (C), METEOR (M), ROUGE-L (R), and SPICE (S) scores per sentence for short descriptions in TACoS Multi-Level Corpus. The best numerical results are marked in bold.}
	\begin{center}
		\begin{tabular}{|M{3.32cm}|M{0.48cm}|M{0.62cm}|M{0.62cm}|M{0.62cm}|M{0.62cm}|}
			\hline
			\textbf{Method}                             & \textbf{B} &  \textbf{C}  & \textbf{M}  &  \textbf{R} & \textbf{S} \\
			\hline \hline
			\begin{tabular}[c]{@{}c@{}}SR Based \citep{rohrbach2014coherent}  \end{tabular}    & 24.7   & - & -   & - & -                                                                                            \\ 
			\begin{tabular}[c]{@{}c@{}}Hierarchical RNN\citep{yu2016video}  \end{tabular}     & 30.5    & 1.602 & 0.287         & - & -                                                                                                                                                \\ 
			\begin{tabular}[c]{@{}c@{}}\textbf{Proposed (Observed)} \end{tabular}  & \textbf{47.6}    & \textbf{1.774} & \textbf{0.352}       & \textbf{0.725} & \textbf{0.162}                         \\
			\hline \hline
			\begin{tabular}[c]{@{}c@{}}Baseline (Unobserved) \end{tabular}  & 10.4   & 0.821 & 0.194      & 0.319 & 0.083       \\
			\begin{tabular}[c]{@{}c@{}}\textbf{Proposed (Unobserved)} \end{tabular}  & \textbf{39.2}    & \textbf{1.493} & \textbf{0.302}       & \textbf{0.631} & \textbf{0.132}                         \\
			\hline         
		\end{tabular}
	\end{center}
	\label{tab:Table6}
\end{table}
\begin{table}[h!]
	\caption{Sequence captioning performance comparisons for TACoS Multi-Level Corpus in terms of BLEU@$4$ (in percent) (B), CIDEr (C), METEOR (M), ROUGE-L (R), and SPICE (S) scores. The best numerical results are marked in bold.}
	\begin{center}
		\begin{tabular}{|M{2.5cm}|M{0.6cm}|M{0.68cm}|M{0.68cm}|M{0.70cm}|M{0.70cm}|}
			\hline
			\textbf{Method} & \textbf{B} & \textbf{C} & \textbf{M} & \textbf{R} & \textbf{S} \\ 
			\hline \hline
			Multi-step Captioning & \multirow{2}{*}{29.9} & \multirow{2}{*}{0.560} & \multirow{2}{*}{0.274} & \multirow{2}{*}{0.609} & \multirow{2}{*}{0.086} \\ 
			(Next-to-Next)  &   &    &   &  & \\ \hline
			\textbf{Proposed Method} & \multirow{2}{*}{	\textbf{30.2}} & \multirow{2}{*}{	\textbf{0.588}} & \multirow{2}{*}{	\textbf{0.291}} & \multirow{2}{*}{	\textbf{0.614}} & \multirow{2}{*}{	\textbf{0.112}} \\ 
			\textbf{(Next-to-Next)}  & &  & & & \\ \hline \hline
			Multi-step Captioning                       &    \multirow{2}{*}{19.8}                                                               &   \multirow{2}{*}{0.548} &   \multirow{2}{*}{0.254}     &       \multirow{2}{*}{0.593} &       \multirow{2}{*}{0.079}                                                                 \\ 
			(Next-to-Next-to-Next)                     &                                                                &      &                  &   &      \\ \hline
				\textbf{Proposed Method}           &      \multirow{2}{*}{	\textbf{20.6}}                                                         & \multirow{2}{*}{\textbf{0.557 }}  &       \multirow{2}{*}{	\textbf{0.264}} &       \multirow{2}{*}{	\textbf{0.598}} &       \multirow{2}{*}{	\textbf{0.103}}                                                                    \\  
			\textbf{(Next-to-Next-to-Next)}                         &                                                              &  &                                &  &                                           \\  \hline
			                                                
		\end{tabular}
	\end{center}
	\label{tab:Table7}
\end{table}
{\bf Experimental Setup.}
For experiments on TACoS Multi-Level Corpus, we use the experimental setup (same test split) of \citep{rohrbach2013translating} which has also been used in \citep{rohrbach2014coherent}. This information is provided with MPII-Cooking $2$ Dataset \citep{rohrbach2016recognizing}. We train our network on a NVIDIA Tesla K80 GPU.

{\bf Quantitative Evaluation.}
Comparisons of our video caption generation results on TACoS Multi-Level Corpus for observed activities with the state-of-the-art methods are shown in Table 6. Not all of the metrics are reported in \citep{rohrbach2014coherent, yu2016video} - hence the missing values. Our method outperforms the state-of-the-arts \citep{yu2016video,rohrbach2014coherent} for observed events. Our method outperforms the baseline for the unobserved events which shows that the generated captions are richer in context than the captions simply obtained by the estimated labels and objects. Quantitative comparisons for captioning a future sequence with the baseline multi-step caption prediction method for TACoS Multi-Level Corpus are shown in Table~\ref{tab:Table7}. As the prediction horizon increases, there is a gradual accumulation of error.

\begin{figure*}[t!]
	\centering
	\includegraphics[width=1\linewidth] {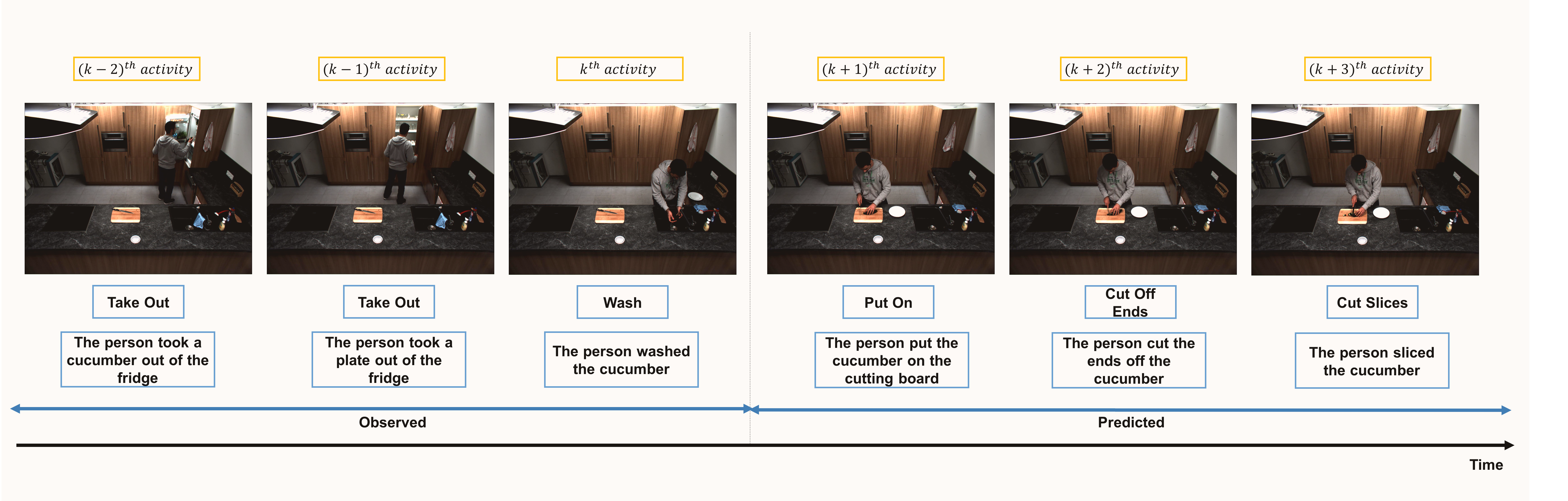}
	\caption{An example activity sequence showing our label prediction and captioning results on TACoS Multi-Level Corpus.}
	\vspace{-2mm}
	\label{fig:fig9}
\end{figure*}

{\bf Qualitative Evaluation.}
Qualitatively, our generated captions like `The person poured the coffee into a cup' is much more meaningful than simple texts like `pour carafe coffee cup' used in the input of the encoder LSTM obtained from the predicted labels and objects. Qualitative Comparisons for captioning a future sequence with the baseline multi-step caption prediction method for TACoS Multi-Level Corpus are shown in Table~\ref{tab:Table8}. Fig.~\ref{fig:fig9} depicts an example sequence showing both of our label prediction and captioning results. 


\begin{table}[!h]
	\caption{Qualitative comparisons of the generated erroneous captions for multi-step caption generation vs. proposed sequential captioning. Mistakes in the captions are marked in bold. Please note that the more we try to predict ahead, the more erroneous the generated captions become.}
	\begin{center}
		\begin{tabular}{|P{1.0cm}|P{2.0cm}|P{2.0cm}|P{2.0cm}|}
			\hline
			\textbf{No. of} &  \textbf{Generated} & \textbf{Generated} & \textbf{Reference} \\
			\textbf{Steps} &  \textbf{Captions} &\textbf{Captions} & \textbf{Captions}\\
			&  \textbf{(Multi-Step)} & \textbf{(Proposed Method)} & \\
			\hline \hline
			2 &The person &The person  &The person\\
			&{\bf sliced} the leek  & peeled the leek  &peeled the leek\\ 
			3 &The person {\bf took out egg} &The person peeled {\bf egg}&The person peeled the leek\\
			
			\hline        
		\end{tabular}
	\end{center}
	\label{tab:Table8}
\end{table} 


\subsubsection{Effect of the Performance of Label Prediction}
We show the BLEU@$4$, CIDEr, METEOR, ROUGE-L, and SPICE scores of our generated captions when generated from the ground truth activity labels and when generated from the predicted labels in Table~\ref{tab:Table9}. The corresponding qualitative comparisons for erroneous results are shown in Table~\ref{tab:Table10}. The type of mistakes made in the generated captions with predicted labels is mostly related to wrong verbs. This is expected since the information regarding the verbs comes from the labels. We get a label prediction accuracy of $65.5\%$ with precision $58.8$ and recall $53.3$ for MPII-Cooking $2$ Dataset \citep{rohrbach2016recognizing} which gives an idea about its effect on the evaluation metrics in Table~\ref{tab:Table9} obtained using ground truth labels and predicted labels.

\begin{table}[!h]
	\caption{Comparisons of BLEU@$4$ (in percent) (B), CIDEr (C), METEOR (M), ROUGE-L (R), and SPICE (S) scores per sentence for short descriptions using ground truth labels vs. predicted labels for caption generation in TACoS Multi-Level Corpus.}
	\begin{center}
		\begin{tabular}{|M{3.0cm}|M{0.5cm}|M{0.64cm}|M{0.64cm}|M{0.64cm}|M{0.64cm}|}
			\hline
			\textbf{Labels Used}                                                            &  \textbf{B} &    \textbf{C} & \textbf{M}  & \textbf{R} & \textbf{S} \\
			\hline \hline
			Ground Truth Labels & 44.0  & 1.615 & 0.351 &0.729 &0.155   \\ 
			Predicted Labels  & 39.2 & 1.493  & 0.302 &0.631 &0.132      \\
			\hline        
		\end{tabular}
	\end{center}
	\label{tab:Table9}
\end{table} 


\begin{table}[!h]
	\caption{Qualitative comparisons of generated erroneous captions using ground truth labels vs. predicted labels for caption generation in TACoS Multi-Level Corpus. Mistakes in the captions are marked in bold.}
	\begin{center}
		\begin{tabular}{|@{\hspace{0.3\tabcolsep}}c@{\hspace{0.3\tabcolsep}}|@{\hspace{0.3\tabcolsep}}c@{\hspace{0.3\tabcolsep}}|@{\hspace{0.3\tabcolsep}}c@{\hspace{0.3\tabcolsep}}|}
			\hline
			\begin{tabular}[c]{@{}c@{}}\textbf{Human} \\ \textbf{Description} \end{tabular} & \begin{tabular}[c]{@{}c@{}}\textbf{Generated Captions} \\\textbf{with Predicted} \\ \textbf{Labels} \end{tabular} & \begin{tabular}[c]{@{}c@{}} \textbf{Generated Captions}  \\ \textbf{with Ground Truth} \\ \textbf{Labels} \end{tabular} \\ \hline \hline
			1. The person sliced &  The person {\bf peeled} &  The person sliced                                                \\
			the carrot     &  the carrot &   the carrot                                                                                                                                                   \\ 
			2. The person  &  The person {\bf cut}  &  The person                                                        \\
			chopped the herbs  &  the herbs &  chopped the herbs                                                           \\
			\hline               
		\end{tabular}
	\end{center}
	\label{tab:Table10}
\end{table} 

\subsubsection{Effect of Scene Context}
MPII-Cooking $2$ Dataset \citep{rohrbach2016recognizing} has many small objects with similar shapes and appearances. Detecting and recognizing these small objects (sometimes with occlusion) in complex videos is a difficult problem itself. The performance of the object recognition method is crucial to the quality of the generated captions. The error of the object recognition method is propagated in two steps: first during label prediction using predicted scene context and then during the mapping from predicted scene context to objects in the captions.  

\begin{table}[t!]
	\caption{Comparisons of BLEU@$4$ (in percent) (B), CIDEr (C), METEOR (M), ROUGE-L (R), and SPICE (S) scores per sentence for short descriptions using ground truth scene context vs. predicted scene context for caption generation in TACoS Multi-Level Corpus.}
	\begin{center}
		\begin{tabular}{|M{3.0cm}|M{0.5cm}|M{0.64cm}|M{0.64cm}|M{0.64cm}|M{0.64cm}|}
			\hline
			\textbf{Scene Context Used}                                                            & \textbf{B} & \textbf{C} & \textbf{M} & \textbf{R} & \textbf{S} \\
			\hline \hline
			Ground Truth  & 39.2 & 1.493 & 0.302 & 0.631 & 0.132      \\ 
			Predicted  & 30.8 & 1.033 & 0.292 & 0.623	 & 0.126    \\
			\hline        
		\end{tabular}
	\end{center}
	\label{tab:Table11}
\end{table} 

\begin{table}[h!]
	\caption{Qualitative comparisons of generated captions using  ground truth scene context vs. using predicted scene context for caption generation in TACoS Multi-Level Corpus. Mistakes in the captions are marked in bold.}
	\vspace{-2mm}
	\begin{center}
		\begin{tabular}{|@{\hspace{0.3\tabcolsep}}c@{\hspace{0.3\tabcolsep}}|@{\hspace{0.3\tabcolsep}}c@{\hspace{0.3\tabcolsep}}|@{\hspace{0.3\tabcolsep}}c@{\hspace{0.3\tabcolsep}}|}
			\hline
			\begin{tabular}[c]{@{}c@{}}\textbf{Human} \\ \textbf{Description} \end{tabular} & \begin{tabular}[c]{@{}c@{}}\textbf{Generated Captions} \\\textbf{with Predicted} \\\textbf{Scene Context} \end{tabular} & \begin{tabular}[c]{@{}c@{}} \textbf{Generated Captions} \\ \textbf{with Ground Truth} \\ \textbf{Scene Context} \end{tabular} \\ \hline \hline
			1. The person cut & The person cut &  The person cut \\
			an orange in half  & the {\bf lime} in half &   {\bf the} orange in half                                                             \\ 
			2. The person took &  The person took &  The person took  \\
			a plum out & a {\bf onion} out &   {\bf plums} out of \\
			of the refrigerator & of the refrigerator &  the refrigerator \\
			\hline        
		\end{tabular}
	\end{center}
	\label{tab:Table12}
\end{table} 


\begin{table}[t!]
	\caption{Comparisons of BLEU@$4$ (in percent) (B), CIDEr (C), METEOR (M), ROUGE-L (R), and SPICE (S) scores per sentence for short descriptions using different length of observed activity sequences in TACoS Multi-Level Corpus. The best numerical results are marked in bold.}
	\vspace{-2mm}
	\begin{center}
	 \begin{tabular}{|c|c|c|c|c|c|}
	 	\hline
		\begin{tabular}[c]{@{}l@{}}\textbf{Observed}\\ \textbf{Sequence}\\ \textbf{Length}\end{tabular} & \textbf{B} & \textbf{C} & \textbf{M} & \textbf{R} & \textbf{S} \\  	\hline 	\hline
		\textbf{2}    & 22.0       &    1.117        &  0.257         &    0.584        & 0.104     \\
		\textbf{3}    &    \textbf{39.2}           &   \textbf{1.493}         &  \textbf{0.302}          &  \textbf{0.631}          & \textbf{0.132}           \\
		\textbf{5}    &     24.0       &  1.078          & 0.262      &   0.597         &   0.107         \\
		\textbf{7}    &   31.9         &   1.142         &     0.284       &  0.614          &    0.117        \\
		\textbf{9}    & 38.5      & 1.156           &  0.297          &   0.628         & 0.128     \\
		\hline
	 \end{tabular}
	\end{center}
	\label{tab:Table13}
\end{table}


\begin{table*}[t!]
	\caption{Qualitative comparisons of the generated erroneous captions using different length of observed activity sequences in TACoS Multi-Level Corpus. Mistakes in the captions are marked in bold. Please note that in most of these erroneous examples, the verbs are incorrect as a result of incorrectly predicted labels.}
	\vspace{-2mm}
	\begin{center}
		\begin{tabular}{|c|c|c|c|}
			\hline
			\textbf{Obs. Seq.} &  \textbf{Predicted}  & \textbf{Generated} & \textbf{Reference} \\
			\textbf{Length} & \textbf{Labels} &  \textbf{Captions} & \textbf{Captions} \\ 
			\hline \hline
			2 & cut apart & \hspace{2mm}The person {\bf cut apart} the leek \hspace{2mm}& The person peeled the leek\\ 
			3 & peel & The person peeled  the leek & The person peeled  the leek\\ 
			5 & slice & The person {\bf sliced} the leek & The person peeled the leek\\ 
			7 & peel & The person peeled the leek & The person peeled the leek \\ 
			9 & slice & The person {\bf sliced} the leek & The person peeled the leek\\
			\hline        
		\end{tabular}
	\end{center}
	\label{tab:Table14}
\end{table*} 

Using the predicted scene context obtained by the object recognition method used in \citep{rohrbach2016recognizing} (combining  dense trajectories, hand trajectories and hand cSift features), we compute the BLEU@$4$, CIDEr, METEOR, ROUGE-L, and SPICE scores for TACoS Multi-Level Corpus. We show the evaluation metrics of our generated captions when generated from the ground truth scene context and from the predicted scene context using the above-mentioned object recognition method in Table~\ref{tab:Table11}. Please note that the evaluation metrics using our caption generation method with predicted scene context is higher than that of the compared methods as well. The corresponding qualitative comparisons for erroneous results are shown in Table~\ref{tab:Table12}. The type of mistakes made in the generated captions with predicted scene context is mostly related to wrong objects. This is expected since the information regarding the objects comes from the scene context. The mean AP using the above-mentioned object recognition method for MPII-Cooking $2$ Dataset \citep{rohrbach2016recognizing} is $43.7\%$  which gives an idea about the relation between the performance of the object recognition method and the performance of label prediction. This in turn shows the effect of scene context on the performance of caption generation. A better object recognition method will lead to better captioning performance.

\subsubsection{Analysis of Observation Horizon}
We empirically find that a sequence length of $3$ for preceding activity features provides best accuracy for label prediction in MPII-Cooking 2 Dataset \citep{rohrbach2016recognizing}. While working with TACoS Multi-Level Corpus, we use observed sequence lengths of $2$, $3$, $5$, $7$ and $9$ and achieved the highest BLEU@$4$, CIDEr, METEOR, ROUGE-L, and SPICE scores for caption generation in TACoS Multi-Level Corpus \citep{rohrbach2014coherent} with an observed sequence length of $3$ as shown in Table~\ref{tab:Table13}. The corresponding qualitative analysis is given in Table~\ref{tab:Table14}. Although the number of wrong words in each sentence is similar, there is reasonable difference in the values of the evaluation metrics in Table~\ref{tab:Table13}. This is because the label prediction performance changes as we change the observed sequence length and this in turn changes the number of such erroneously generated captions.

\section{Conclusions}
\label{section:Conclusion}
In this work, we proposed a solution to a novel problem of predicting the labels and captions of a sequence of future unobserved activities. We took advantage of the combination of LSTM and fully connected layers to exploit the contextual relationship among activities and objects for label prediction. For mapping the predicted labels and scene context to meaningful captions, we incorporated a sequence-to-sequence learning-based approach using an encoder-decoder LSTM pair. Rigorous experimental analysis on challenging datasets proves the robustness of our framework. In the future, we plan to extend our prediction method to a multi-camera environment and investigate how to predict new unseen activity classes. Another interesting direction for future work could be tackling the domain adaptation problem where we predict the labels and the captions of a dataset by training on the labels and the captions of a different dataset.

\section{Acknowledgments}
This work was partially supported by NSF grant 1724341 and ONR grant N00014-19-1-2264. We gratefully acknowledge the support of NVIDIA with the donation of the Tesla K80 GPU used for this research.



%

\bibliographystyle{elsarticle-num}
\bibliography{tip_prediction}

\end{document}